\documentclass[conference]{IEEEtran}
\IEEEoverridecommandlockouts
\usepackage{cite}
\usepackage{amsmath,amssymb,amsfonts}
\usepackage{algorithmic}
\usepackage{graphicx}
\usepackage{textcomp}
\usepackage{xcolor}
\usepackage{multirow} % 表格多行合并
\usepackage{bbding} % 对号
\usepackage{utfsym} % 错号
\usepackage{pifont} % 错号
\usepackage{makecell} % 单元格内换行
\usepackage{graphicx} % 表格调整

\def\BibTeX{{\rm B\kern-.05em{\sc i\kern-.025em b}\kern-.08em
    T\kern-.1667em\lower.7ex\hbox{E}\kern-.125emX}}

\begin{document}

\title{MMHMER: Multi-viewer and Multi-task for Handwritten Mathematical Expression Recognition}

\author{
\IEEEauthorblockN{1\textsuperscript{st} Kehua, Chen, 2\textsuperscript{nd}  Haoyang, Shen, 3\textsuperscript{rd} Lifan, Zhong, \\
4\textsuperscript{th} Mingyi, Chen}
ShenZhen, China\\
\{chenkehua880@gmail.com, shenhaoyang6@gmail.com, zhong\_lifan@yahoo.co.jp, zhalandingli@gmail.com}

\maketitle

\begin{abstract}
Handwritten Mathematical Expression Recognition (HMER) methods have made remarkable progress, with most existing HMER approaches based on either a hybrid CNN/RNN-based with GRU architecture or Transformer architectures. Each of these has its strengths and weaknesses. Leveraging different model structures as viewers and effectively integrating their diverse capabilities presents an intriguing avenue for exploration. This involves addressing two key challenges: 1) How to fuse these two methods effectively, and 2) How to achieve higher performance under an appropriate level of complexity. This paper proposes an efficient CNN-Transformer multi-viewer, multi-task approach to enhance the model's recognition performance. Our MMHMER model achieves 63.96\%, 62.51\%, and 65.46\% ExpRate on CROHME14, CROHME16, and CROHME19, outperforming Posformer with an absolute gain of 1.28\%, 1.48\%, and 0.58\%. The main contribution of our approach is that we propose a new multi-view, multi-task framework that can effectively integrate the strengths of CNN and Transformer. By leveraging the feature extraction capabilities of CNN and the sequence modeling capabilities of Transformer, our model can better handle the complexity of handwritten mathematical expressions. 
\end{abstract}

\begin{IEEEkeywords}
HMER, Multi-Viewer, Multi-Task, CNN/RNN, Transformer
\end{IEEEkeywords}

\section{Introduction}
Handwritten Mathematical Expression Recognition (HMER) plays a crucial role in the field of document analysis. Traditional HMER methods typically employ a three-stage approach: a symbol segmentation step, a symbol recognition step, and a grammar-guided structure analysis step. In the recognition step, classical classification techniques such as HMM \cite{b1,b2,b24}, Elastic Matching\cite{b3,b25,b26}, and Support Vector Machines(SVM)\cite{b4} are primarily used. Traditional methods are insufficient in practical applications due to their limited feature-learning capabilities and complex grammar rules. With the widespread application of machine learning and deep learning networks in academia and their impressive performance, Encoder-Decoder architectures are increasingly being applied to handwritten expression recognition. This end-to-end recognition framework, similar to global recognition methods, significantly improves the accuracy of handwritten expression recognition and effectively handles strong contextual dependencies within expression. Within the encoder-decoder framework, the encoder typically uses DenseNet\cite{b5} for feature extraction, while the decoder usually employs tree-based structures using CNN and RNN with Gated Recurrent Units (GRU) \cite{b27} or Transformer \cite{b30} to parse the expression structure.

CNNs have the ability to model locally, while Transformers can model non-local information. These two methods each have their own distinct advantages. Exploring whether the strengths of these two methods can be effectively combined to enhance HMER performance in the image domain is valuable. The work LICTCM \cite{b6} proposes an efficient parallel Transformer-CNN hybrid block that combines the advantages of both Transformers and CNNs, which significantly improves the performance of the distortion rate and the operational efficiency of image compression through an innovative parameter-efficient attention module and an optimized entropy model design. The work \cite{b7} introduces a novel convolutional module combined with a transformer to enhance object detection accuracy by fusing detailed and global features, achieving a 1.7\% mAP improvement on YOLOv5n and outperforming faster RCNN\cite{b32} with 81\% accuracy using fewer parameters on the Pascal VOC dataset\cite{b31}. A good student \cite{b8} proposes an online knowledge distillation strategy that achieves collaborative learning of CNN and ViT models for semantic segmentation through heterogeneous feature distillation and bidirectional selective distillation, promoting their complementary learning of global and local feature representations; the method achieves new state-of-the-art performance on benchmark datasets, demonstrating its effectiveness in pushing the limits of semantic segmentation technology. In the field of HMER, there has been little work exploring how to effectively utilize both CNNs and transformers to enhance performance. Another work \cite{b9} investigates three attention-based encoder-decoder (AED) models that combine Transformers and CNNs for handwritten mathematical expression recognition, discovering that the series and parallel approaches outperform the mixing approach on the CROHME benchmark, with both showing similar performance. The experimental results indicate that using Transformers to capture global dependencies can effectively improve recognition rates. 

In this paper, we strive to propose a framework which collaboratively learn compact and effective CNN-based and transformer-based models to explore the boundaries of model capabilities. Intuitively, we adopt a “Dual-Student” framework, using different student models as different observers, to collaboratively learn under different task identities in order to enhance the robustness of the backbone structure. However, these multi-task learning methods are less effective and can even lead to performance degradation. The reasons are: 1) The gap in model size and learning capacity between CNNs and transformers. 2) The differences in prediction spaces between CNNs and transformers due to different computational paradigms, posing challenges for online learning of the backbone. 3) Excessive task differences may confuse the feature extracting of model. In light of this, we propose the first online multi-perspective multi-task learning method to date, to further push the limits of CNNs and transformers in HMER. Our method has two breakthroughs. 

    1. We propose an online collaborative learning strategy where CNN and Transformer act as different observers, focusing on the local structure and global context understanding of handwritten mathematical expressions, respectively. 
    
    2. We consider not only the cross-entropy guidance between predictions and true labels but also encourage the two observers to compensate for each other’s shortcomings through consistency constraints, thus achieving finer model optimization in the HMER task. 

\section{RELATED WORKS}
\subsection{Mixed CNN/RNN-based Models with GRU}
The work \cite{b14} encodes the input two-dimensional stroke trajectory information of handwritten expressions using a Gated Recurrent Unit-based Recurrent Neural Network (GRU-RNN). The decoder is also implemented with GRU-RNN and equipped with a coverage-based attention model. The effectiveness of the recognition process is demonstrated through the attention mechanism. TAP \cite{b15} architecture consists of two main components: the Tracker and the Parser. The Tracker utilizes a stacked bidirectional recurrent neural network with Gated Recurrent Units (GRUs) to model the handwriting trajectory, thereby fully leveraging the dynamic information during the handwriting process. The Parser, on the other hand, employs a GRU with a Guided Hybrid Attention (GHA) mechanism to generate mathematical symbols.  "Watch, attend and parse" \cite{b10} employs a fully convolutional encoder to extract image features and introduces a coverage attention mechanism to address inaccuracies in attention during long-distance decoding, thereby establishing the WAP model as a foundational model for most sequence-based recognition methods. The research \cite{b11} further improved upon this by proposing DenseWAP, which adopts a DenseNet structure to enhance the encoder and introduces a multi-scale attention decoding model to address the challenge of recognizing characters of different sizes in formulas. CAN \cite{b12} proposes a method that utilizes symbol counting as an auxiliary task to enhance the robustness of the encoder-decoder model in HMER, improving recognition accuracy by providing symbol-level positional information and global counting information. SAM \cite{b13} constructed a semantic graph based on statistical co-occurrence probabilities, explicitly showing the dependencies between different symbols; it also proposed a semantic-aware module that takes visual and classification features as input and maps them into semantic space. SemiHMER \cite{b33} uses two parallel CNN networks with consistency regularization, pseudo labeling, a weak to strong augmentation schedule, and a Global Dynamic Counting Module to leverage unlabeled data and reduce long distance and repeated symbol errors in existing HMER encoder and decoder models.

\subsection{Transformer-based Models}
Existing RNN-based encoder-decoder models face issues of insufficient coverage in the task of HMER, manifesting as over-parsing and under-parsing, and struggling with handling long-distance symbol relationships, especially when parsing LaTeX language. BTTR \cite{b16} introduces a transformer decoder that utilizes positional encoding to improve coverage problems and proposes a bidirectional training strategy, enabling a single decoder to perform both left-to-right and right-to-left decoding simultaneously. This enhances the model’s training efficiency and inference performance, outperforming RNN-based models. Due to the lack of a coverage attention mechanism, CoMER\cite{b17} proposes a new model that introduces an Attention Refinement Module (ARM) and self-coverage and cross-coverage mechanisms, which address the shortcomings of the Transformer decoder in handling coverage attention. As a result, CoMER achieves better performance in the HMER task compared to the standard Transformer and RNN decoders. TSDNet \cite{b28} designed a transformer-based tree decoder that can better capture complex correlations. PosFormer \cite{b18} introduces a novel Position Forest Transformer, which enhances the positioning understanding of sequence-based methods by encoding expressions into a forest structure and parsing their nested hierarchies and relative positions. This method aids expression recognition through a position identification task and incorporates an implicit attention correction module to improve the accuracy of the decoder. PosFormer has demonstrated outstanding performance on multiple benchmark datasets without incurring additional latency or computational costs. NAMER \cite{b29} proposes a novel non-autoregressive modeling approach for handwritten mathematical expression recognition that significantly outperforms state-of-the-art methods in accuracy and decoding speed. 

\section{METHODOLOGY}

\begin{figure*}[htbp]
    \centering
    \includegraphics[width=0.7\linewidth]{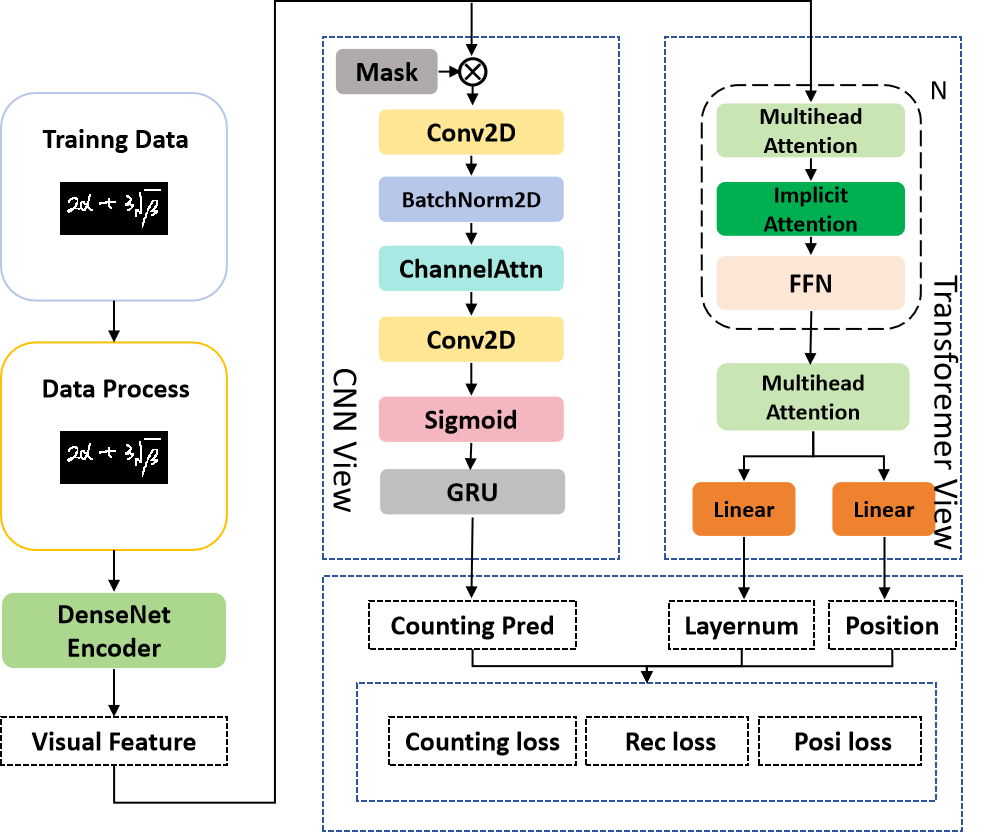}
    \caption{Overview of multi-viewer and multi-task for Handwritten Mathematical Expression Recognition}
    \label{fig1}
\end{figure*}

In this section, we will describe the proposed Handwritten Mathematical Expression Recognition framework, namely the Convolutional Neural Network (CNN) and Transformer multi-viewer and multi-task network, as illustrated in Figure 1. The proposed framework consists of two parallel branches: the Transformer prediction branch and the CNN prediction branch. We initially employ the DenseNet backbone network to extract high-level feature representations. In the Transformer prediction branch, multiple Multihead-Attention modules, Implicit Attention modules, and FFN modules are used to efficiently parse formula symbols.  Finally, a position forest and an expression recognition head, along with an attention-based transformer decoder, are utilized to obtain discriminative symbol features. In addition to expression recognition, the position forest is introduced for joint optimization to facilitate the learning of position-aware symbol-level feature representations. In the CNN prediction branch, a GRU is used to parse the formula, and a multi-scale counting module is employed to predict the number of each symbol class and generate representations of the counting results.

\subsection{Transformer viewer}
The Transformer model has achieved unprecedented accuracy in recognizing complex and nested handwritten mathematical expressions. This accomplishment is made possible by the innovative integration of positional forests and an implicit attention correction module. The structural design of the positional forest allows the model to precisely grasp the interrelations between symbols and their specific positions within the overall expression, greatly enhancing the model’s deep understanding of the structural characteristics of mathematical expressions. Meanwhile, the implicit attention correction module ensures that the model can focus on the key parts of the expression while processing information, effectively identifying and correcting any errors that may have arisen during the encoding phase. The synergistic effect of these two modules not only improves the accuracy of recognition but also enhances the model’s ability to parse complex mathematical expressions, bringing about significant technological advancements in the field of handwritten mathematical expression recognition.

The process of encoding handwritten mathematical expressions using the Transformer begins with subdividing the entire expression into smaller substructures. These substructures are diverse, including not only superscript-subscript structures, fraction structures, and square root structures, but also special operator structures, and more. After the expression is broken down into these substructures, we proceed to classify the symbols within each substructure based on their relative positional relationships. This step involves categorizing each symbol as “upper,” “lower,” or “middle.” Subsequently, these substructures are encoded in a left-to-right order, with each substructure being transformed into a tree-like structure. In these trees, the main part of the substructure serves as the root node, the upper part as the left child node, and the lower part as the right child node. Ultimately, the encoded tree structures are arranged or nested in sequence to form a positional forest structure, which not only captures the complex relationships between substructures but also enables the model to accurately recognize and parse handwritten mathematical expressions.

The use of the positional forest, combined with the implicit attention correction module, greatly expands the Transformer’s capability to handle mathematical expressions. It can easily cope with simple arithmetic operations and also parse equations involving multiple variables and complex operators. The model’s deep understanding of the structural attributes of mathematical expressions allows it to maintain good generalization capabilities when dealing with unseen data. This feature makes the Transformer a powerful tool for recognizing and parsing handwritten mathematical expressions, showing its immense potential and application value in fields such as education, scientific research, and any other domain that requires the processing of mathematical symbols.

\subsection{CNN Viewer with GRU}

Following the methodology outlined in CAN \cite{b12}, the Multi-Scale Counting Module (MSCM) emerges as a critical component within the model's architecture in CNN Viewer, meticulously crafted to predict the frequency of occurrence for each symbol class within the visual representation of a mathematical expressions. The conception of this module is deeply rooted in the necessity to tackle the pronounced variability in symbol sizes and styles that is characteristic of handwritten mathematical expressions—a challenge that conventional single-scale methods are inadequately equipped to address with precision.

The initial stage of the MSCM involves extracting features from the input image. This is accomplished through the use of two parallel convolutional branches, each equipped with distinct kernel sizes. Specifically, the kernels are set at dimensions of 3x3 and 5x5. This dual-kernel approach plays a crucial role in capturing a wide range of features at various scales, thereby enhancing the model's ability to distinguish both minute details and broader structural elements within the formula. The logic behind this multi-scale feature extraction is to emulate the human visual system's capability to perceive objects at various distances and sizes, which is particularly vital when dealing with mathematical notations that can vary from tiny subscripts to large integral signs.

Following feature extraction, the MSCM integrates channel attention mechanisms. Channel attention is an advanced technique that dynamically adjusts the model's focus on different feature channels, allowing it to highlight those most relevant to the current task. This selective enhancement of feature information not only aids in improving the accuracy of symbol recognition but also helps in reducing noise and irrelevant data, which is often present in handwritten or printed mathematical expressions due to variations in writing styles or print quality.

The final operation within the MSCM involves applying sum pooling to the resulting feature maps from the convolutional and channel attention layers. Sum pooling, as opposed to other pooling methods like max or average pooling, aggregates feature values across spatial dimensions by summing them. This method is selected for its simplicity and effectiveness in reducing the dimensionality of feature maps while retaining essential information. By doing so, sum pooling not only lightens the computational burden on subsequent layers but also makes the model more robust against minor spatial shifts or distortions in the input image, which are common in real-world mathematical formula images.

\subsection{Loss Function}
Given a handwritten mathematical expression image $\mathbf{X}\in \mathbb{R}^{H \times W}$, our objective is to interpret the expression and convert it into a corresponding character sequence $\mathcal{Y}_{c}=\big\{y_{c}^{(t)}|y_{c}^{(t)} \in \{$``\texttt{a}'', ``\texttt{b}'', $\cdots$, ``\texttt{\_}''$\}\big\}_{t=1}^{T}$. This sequence $\mathcal{Y}_{c}$ consists of characters where each character is drawn from a predefined set "a", "b", ... "\_", and the sequence has a length of T. 
H, W, and T are the image height, image width respectively.
\begin{equation}
    \mathcal{L}_{\rm rec} = -\frac{1}{T}\sum_{t=1}^{T} y_{c}^{(t)} \log p(y_{c}^{(t)})
    \label{Eq1}
\end{equation}
The transformer viewer component is trained in an end-to-end manner within a multi-task framework, to optimize a unified objective function that simultaneously addresses multiple tasks. 
 \begin{equation}
    \mathcal{L}_{\rm pos} = -\frac{1}{T}\sum_{t=1}^{T} \big(y_{n}^{(t)} \log p(y_{n}^{(t)}) + y_{r}^{(t)} \log p(y_{r}^{(t)})\big)
        \label{Eq2}
\end{equation}
 The sequence $\mathcal{Y}_{c}$ denotes the groundtruth LaTeX sequence, which is
 \begin{equation}
    \mathcal{Y}_{c}=\{y_{c}^{(t)}\}_{t=1}^{T}
        \label{Eq3}
\end{equation}
$\mathcal{Y}_{n}$ indicates the real nesting depth levels, which is,
 \begin{equation}
    \mathcal{Y}_{n}=\{y_{n}^{(t)}\}_{t=1}^{T}
        \label{Eq4}
\end{equation}
$\mathcal{Y}_{r}$ signifies the actual relative positions, which is,
 \begin{equation}
   \mathcal{Y}_{r}=\{y_{r}^{(t)}\}_{t=1}^{T}
        \label{Eq5}
\end{equation}
The $p(y_{c}^{(t)})$, $p(y_{n}^{(t)})$, and $p(y_{r}^{(t)})$ denotes the predicted distributions of three tasks.
The CNN viewer part is denoted the counting ground truth of each symbol class as $\hat{\mathcal{V}}$, $\mathcal{L}_{counting}$ is a smooth $L1$ \cite{b22} regression loss defined as follows:
\begin{equation}
\label{eq6}
{\mathcal{L}_{counting}=smooth_{L1}(\mathcal{V}, \hat{\mathcal{V}})}
\end{equation}

Finally, the overall training loss is summarized as:
\begin{equation}
    \begin{aligned}
    \mathcal{L}_{\rm all} = \lambda_{1} \cdot  \mathcal{L}_{\rm rec}  + \lambda_{2} \cdot  \mathcal{L}_{\rm pos} + \lambda_{3}\cdot \mathcal{L}_{counting}
    \label{Eq7}
    \end{aligned}
  \end{equation}
where $\lambda_{1}$ , $\lambda_{2}$ and $\lambda_{3}$ are a loss coefficient. $\lambda_{1}$ default is 1 and the default value of $\lambda_{2}$ is set to 0.5. $\mathcal{L}_{\rm rec}$ represents the loss for character recognition, $\mathcal{L}_{\rm pos}$ represents the loss for character position.
  
\section{Experiment}
\subsection{Datasets}

\begin{figure}
    \centering
    \includegraphics[width=0.95\linewidth]{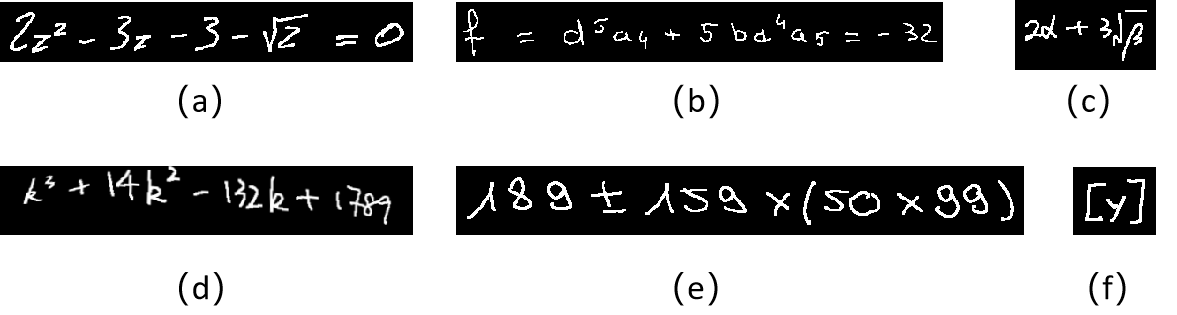}
    \caption{Sample Images from CROHME 2014/2016/2019 Datasets}
    \label{fig2}
\end{figure}

\textbf{CROHME Dataset} is the most widely used public dataset in the HMER field, originating from the online handwritten mathematical expression recognition competition CROHME. The CROHME training set contains 8836 handwritten mathematical expressions, and there are three test sets: CROHME 2014\cite{b19}, 2016\cite{b20}, and 2019\cite{b21}, which include 986, 1147, and 1199 handwritten mathematical expressions, respectively. They have 110 symbol classes, excluding sos" and "eos". In the CROHME dataset, each handwritten mathematical expression is stored in text format, which records the trajectory coordinates of the handwritten strokes. 

\textbf{MNE Dataset} is introduced in the Posformer paper \cite{b18}, is a meticulously designed Multi-level Nested Handwritten Mathematical Expression test set. Its primary purpose is to assess the efficacy of the model in recognizing intricate mathematical expression images. This dataset is structured into three distinct subsets: N1, N2, and N3, each corresponding to different levels of nesting 1, 2, and 3, respectively. The subsets contain a total of 1875 images for N1, 304 images for N2, and 1464 images for N3. This composition underscores the importance of enhanced position-aware symbol feature extraction in the context of recognizing nested mathematical expressions. The varying levels of complexity within the MNE Dataset provide a comprehensive framework for evaluating the model’s performance across a range of challenges, thereby ensuring a robust assessment of its capabilities in the field of handwritten mathematical expression recognition.

\subsection{Metrics}
Expression Recognition Rates are widely used metrics in the field of HMER to measure performance. "$\leq 1$", "$\leq 2$", and "$\leq 3$" represent the rate of expression recognition when tolerating from 0 to 3 symbol-level errors.

\subsection{Implementation Details}
Following the Posformer\cite{b18} approach, the DenseNet architecture was used as the backbone network in the encoder. In the Transformer decoder, we employ a three-layer decoder as the position forest decoder, along with a linear layer for symbol classification. For the position recognition task, we use two linear layers with different output dimensions. In multi-line formulas, each line is transformed into multiple trees corresponding to the substructures, and the trees from different lines are arranged sequentially to form a forest. In the CNN decoder, we follow the CAN\cite{b12} method and adopt a Counting and Attention-based Decoder (CCAD), using 1 $\times 1$ convolutions to change the number of channels. After obtaining the transformed features, we apply a fixed absolute position encoding. Our code is implemented using PyTorch, and we train our model on a Nvidia Tesla V100 graphics card with 32GB of memory. We adhere to the same training parameters as Posformer, including batch size, learning rate, and optimizer. 

\subsection{Comparison with State-of-the-Art Methods}

\textbf{Results on the CROHME Dataset.} We present the outcomes of our MMHMER model in comparison to the existing State-of-the-Art (SOTA) approaches. Table \ref{tab1} encapsulates the comparative performance metrics. We have taken care to ensure that the comparison is equitable by including the results of MMHMER both with and without the application of data augmentation techniques. In the scenarios where data augmentation is not utilized, MMHMER demonstrates its superiority by outperforming the previous SOTA results with incremental gains of 0.4\% on the CROHME 2014 test set, 0.43\% on the CROHME 2016 test set and 0.33\% on the CROHME 2019 test set. When data augmentation is incorporated into the training process, MMHMER’s performance edge becomes even more pronounced, with significant enhancements in the ExpRate metric of 1.28\% for the 2014 set, 1.48\% for the 2016 set, and 0.58\% for the 2019 set, thereby demonstrating the model’s effectiveness and the benefits of data augmentation in this context.

\begin{table}[htbp]
\centering
\tabcolsep=2pt
\renewcommand\arraystretch{1.2}
\caption{\centering Performance comparison with previous SOTA methods on the CROHME 2014, 2016, and 2019 test sets (in \%). The notation "Scale-aug" denotes scale augmentation.}
    \begin{tabular}{c|c|c|c|ccc}
    \hline 
        Dataset & Scale-aug & Model & ExpRate $\uparrow$ & $\leq 1 \uparrow$ & $\leq 2 \uparrow$ & $\leq 3 \uparrow $\\
    \hline 
        \multirow{8}{*}{CROHME14}& \multirow{5}{*}{\textcolor{red}{\ding{55}}}& DWAP& 50.10 & - & - & - \\
        \multirow{8}{*}{}& \multirow{5}{*}{}& BTTR & 53.96 & 66.02 & 70.28 & - \\
        \multirow{8}{*}{}& \multirow{5}{*}{}& CAN& 57.26 & 74.52 & 82.03 & - \\
        \multirow{8}{*}{}& \multirow{5}{*}{}& PosFormer & 60.45 & 77.28 & \textbf{83.68}& \textbf{87.83}\\
 \multirow{8}{*}{}& \multirow{5}{*}{}& MMHMER& \textbf{60.85(+0.4)}& \textbf{77.38}& 83.37&87.53\\
    \cline{2-7}
        \multirow{8}{*}{}& \multirow{4}{*}{\textcolor{green}{\Checkmark}}& CoMER & 59.33 & 71.70 & 75.66 & 77.89 \\
        \multirow{8}{*}{}& \multirow{4}{*}{}& BPD-Coverage & 60.65 & - & - & - \\
        \multirow{8}{*}{}& \multirow{4}{*}{}& PosFormer & 62.68 & \textbf{79.01}& 84.69 & \textbf{88.84}\\
 \multirow{8}{*}{}& \multirow{4}{*}{}& MMHMER& \textbf{63.96(+1.28)}& 78.68& \textbf{84.87}&88.32\\
    \hline
        \multirow{8}{*}{CROHME16}& \multirow{5}{*}{\textcolor{red}{\ding{55}}}& DWAP & 47.50 & - & - & - \\
        \multirow{8}{*}{}& \multirow{5}{*}{}& BTTR & 52.31 & 63.90 & 68.61 & - \\
        \multirow{8}{*}{}& \multirow{5}{*}{}& CAN & 56.15 & 72.71 & 80.30 & - \\
        \multirow{8}{*}{}& \multirow{5}{*}{}& PosFormer & 60.94 & \textbf{76.72}& \textbf{83.87}& 88.06 \\
 \multirow{8}{*}{}& \multirow{5}{*}{}& MMHMER& \textbf{61.37(+0.43)}& 76.37& 83.44&\textbf{88.49}\\
    \cline{2-7}
        \multirow{8}{*}{}& \multirow{4}{*}{\textcolor{green}{\Checkmark}}& CoMER & 59.81 & 74.37 & 80.30 & 82.56 \\
        \multirow{8}{*}{}& \multirow{4}{*}{}& BPD-Coverage & 58.50 & - & - & - \\
        \multirow{8}{*}{}& \multirow{4}{*}{}& PosFormer & 61.03 & \textbf{77.86}& \textbf{84.74}&89.28\\
 \multirow{8}{*}{}& \multirow{4}{*}{}& MMHMER& \textbf{62.51(+1.48)}& 77.68& 84.31&\textbf{88.92}\\
    \hline
        \multirow{7}{*}{CROHME19}& \multirow{4}{*}{\textcolor{red}{\ding{55}}}& BTTR & 52.96 & 65.97 & 69.14 &- \\
        \multirow{7}{*}{}& \multirow{4}{*}{}& CAN & 55.96 & 72.73 & 80.57 & - \\
        \multirow{7}{*}{}& \multirow{4}{*}{}& PosFormer & 62.22 & \textbf{79.40}& \textbf{86.57}& 89.09 \\
 \multirow{7}{*}{}& \multirow{4}{*}{}& MMHMER& \textbf{62.55(+0.33)}& 79.23& 86.32&\textbf{89.49}\\
     \cline{2-7}
        \multirow{7}{*}{}& \multirow{4}{*}{\textcolor{green}{\checkmark}}& CoMER & 62.97 & 77.40 & 81.40 & 83.07\\
        \multirow{7}{*}{}& \multirow{4}{*}{}& BPD-Coverage & 61.47 & - & - & - \\
 \multirow{7}{*}{}& \multirow{4}{*}{}& PosFormer& 64.97& \textbf{82.49}& 87.24&87.24\\
        \multirow{7}{*}{}& \multirow{4}{*}{}& MMHMER& \textbf{65.55(+0.58)}& 82.15& \textbf{87.82}& \textbf{91.57}\\
    \hline
    \end{tabular}
\label{tab1}
\end{table}

\textbf{Results on the MNE Dataset.} In Table \ref{tab2}, we conducted tests on the N1, N2, and N3 subsets, and the results showed performance improvements of 1.44\%, 0.65\%, and 0.21\% on each subset, respectively. This indicates that the performance provided by MMHMER is further enhanced, highlighting the importance of the multi-viewer and multi-task approach for improving the model’s effectiveness. 

\begin{table}[htbp]
\centering
\tabcolsep=2pt
\renewcommand\arraystretch{1.4}
\caption{\centering Performance comparison with previous SOTA methods on the complex MNE test set, with ExpRate, $\leq 1$, $\leq 2$, and $\leq 3$ values shown in percentages(\%).}
    \begin{tabular}{c|c|c|c|ccc}
    \hline 
        Dataset & Size& Model & ExpRate $ \uparrow $ & $\leq 1 \uparrow $ & $\leq 2 \uparrow $ & $\leq 3 \uparrow $\\
    \hline 
        \multirow{3}{*}{N1}& \multirow{3}{*}{1875}& COMER& 59.73& 77.55& 84.11& 88.91\\
        \multirow{3}{*}{}& \multirow{3}{*}{}& PosFormer& 60.59& 77.97& \textbf{84.32}& \textbf{88.75}\\
 \multirow{3}{*}{}& \multirow{3}{*}{}& MMHMER& \textbf{62.03(+1.44)}& \textbf{78.13}& 84.32&88.48\\
    \hline
        \multirow{3}{*}{N2}& \multirow{3}{*}{304}& COMER& 37.17& 53.95& 65.13& 72.37\\
        \multirow{3}{*}{}& \multirow{3}{*}{}& PosFormer& 38.82& \textbf{56.91}& \textbf{66.12}& 73.36\\
 \multirow{3}{*}{}& \multirow{3}{*}{}& MMHMER& \textbf{39.47(+0.65)}& 56.57& 65.46&\textbf{73.68}\\
    \hline
        \multirow{3}{*}{N3}& \multirow{3}{*}{1464}& COMER& 24.04& 32.31& 36.34&39.89\\
        \multirow{3}{*}{}& \multirow{3}{*}{}& PosFormer& 34.08& \textbf{36.82}& 40.30& \textbf{43.10}\\
        \multirow{3}{*}{}& \multirow{3}{*}{}& MMHMER& \textbf{34.29(+0.21)}& 36.68& \textbf{40.51}& 42.83\\
    \hline
    \end{tabular}
\label{tab2}
\end{table}
\section*{Ablations and Analysis}

\begin{table}[htbp]
\centering
\tabcolsep=2pt
\renewcommand\arraystretch{1.3}
\caption{\centering Ablation to investigate the impact of Multi-Viewer and Multi-Task on the Performance.}
    \begin{tabular}{c|c|c|cc}
    \hline 
        Dataset & Strategy & ExpRate $ \uparrow $ & $\leq 1 \uparrow $ & $\leq 2 \uparrow $\\
    \hline 
        \multirow{4}{*}{CROHME14}& baseline& 62.68& 79.01& 84.69\\
        \multirow{4}{*}{}& +Task1& 63.05& 78.80& 84.48\\
        \multirow{4}{*}{}& +Task1 in multi-view& 63.96& 78.68& 84.87\\
        \multirow{4}{*}{}& +Task1,Task2 in multi-view& 58.21& 74.95& 81.33\\
    \hline
        \multirow{4}{*}{CROHME16}& baseline
& 61.03& 77.86& 84.74\\
        \multirow{4}{*}{}& +Task1& 60.85& 77.68& 84.13\\
        \multirow{4}{*}{}& +Task1 in multi-view& 62.51& 77.68& 84.31\\
        \multirow{4}{*}{}& +Task1,Task2 in multi-view& 57.11& 74.11& 81.08\\
    \hline
        \multirow{4}{*}{CROHME19}& baseline
& 64.97& 82.49& 87.24\\
        \multirow{4}{*}{}& 

+Task1& 65.13& 82.15& 87.32\\
        \multirow{4}{*}{}& +Task1 in multi-view& 65.55& 82.15& 87.82\\
        \multirow{4}{*}{}& +Task1,Task2 in multi-view& 57.54& 75.06& 82.07\\
    \hline
    \end{tabular}
\label{tab3}
\end{table}

\textbf{Effectiveness of Multi-Viewer.} As illustrated in Table \ref{tab3}, we have showcased the performance enhancements attributed to the multi-viewer component of the MMHMER system. We define the symbol counting task as Task1. When the symbol count is incorporated into the original model and contrasted with previous research, the performance on the CROHME 2014/2016/2019 test sets sees a modest increase of 0.37\%, a decrease of -0.18\%, and an improvement of 0.16\%, respectively. This indicates that while the inclusion of symbol counting does contribute to a slight enhancement in Handwritten Mathematical Expression Recognition (HMER) performance, the impact is relatively minor. However, by integrating a Convolutional Neural Network (CNN) as an additional branch to handle Task1, thereby implementing the multi-viewer strategy, we achieve the current state-of-the-art results. This clearly demonstrates the value and effectiveness of the multi-viewer approach in enhancing the overall performance of the HMER system. The multi-viewer’s ability to leverage different types of features and tasks concurrently leads to a more comprehensive and accurate recognition of handwritten mathematical expressions, highlighting its contribution to the advancement of HMER technology.

\textbf{Effectiveness of Multi-Task.} As we integrate additional tasks into our model, such as text prediction designated as Task2, we notice a marked deterioration in the overall performance metrics. Specifically, the inclusion of this extra task results in a decrease in performance by 5.75\% on the 2014 dataset, 5.40\% on the 2016 dataset, and a more pronounced drop of 8.01\% on the 2019 dataset. These findings suggest that an overload of tasks introduces greater complexity, which in turn complicates the model’s ability to achieve convergence. The added computational burden and the potential for task interference appear to outweigh the benefits of multi-task learning in this context, underscoring the importance of carefully selecting and balancing the number of tasks a model is designed to handle.

\begin{table}[htbp]
\centering
\tabcolsep=2pt
\renewcommand\arraystretch{1.2}
\caption{\centering Ablation to investigate the impact of $\lambda_{3}$ on the Performance.}
    \begin{tabular}{c|c|c|c|c}
    \hline 
        Dataset & $\lambda_{3}$& CROHME14&CROHME16 &CROHME19\\
    \hline 
        \multirow{3}{*}{ExpRate $\uparrow$}& 0.01& 61.87 &60.51 &63.55\\
        \multirow{3}{*}{}& 0.1& 63.96 &62.51 &65.55\\
        \multirow{3}{*}{}& 0.5& 59.63& 59.11&61.80\\
    \hline
    \end{tabular}
\label{tab4}
\end{table}
As shown in Table \ref{tab4}, the symbol $\lambda$ is used to balance the trade-off between tasks for CNNs and transformers in the experiments above. In our setting, $\lambda$=0.1 performs the best, whereas a smaller $\lambda$ would reduce the learning capability of the CNN. On the other hand, a larger $\lambda$ presents issues and can lead to performance degradation because the network might converge in the wrong direction. 

\section*{Conclusion}

In this study, we propose a novel approach for Handwritten Mathematical Expression Recognition (HMER) by leveraging the complementary strengths of Convolutional Neural Networks (CNNs) and Transformer architectures within a multi-viewer and multi-task learning framework. Our primary motivation is to address the inherent challenges of HMER. By integrating CNNs and Transformers, we aim to harness the local feature extraction capabilities of CNNs and the global contextual understanding of Transformers, thereby enhancing the robustness and accuracy of the HMER system.

The multi-view aspect of our framework allows us to process handwritten mathematical expressions through both CNN-based and Transformer-based pathways, capturing diverse representations. The CNN pathway excels in extracting low-level features such as strokes and local patterns, while the Transformer pathway focuses on modeling long-distance dependencies and hierarchical structures within the expressions. This dual-view approach not only improves the model’s ability to handle complex expressions but also provides a more comprehensive understanding of the input data, leading to better generalization across different handwriting styles and expression types.

Furthermore, the multi-task learning paradigm integrated into our framework enables simultaneous optimization of symbol recognition and expression structure prediction. By sharing representations between these tasks, the model is able to learn more meaningful and task-relevant features, thus improving overall performance. Experimental results indicate that our proposed method outperforms baseline methods and state-of-the-art HMER systems, achieving significant improvements in recognition accuracy and structural correctness. In summary, our work highlights the effectiveness of combining CNNs and Transformers in a multi-view and multi-task framework for HMER. The synergy between these architectures not only addresses the unique challenges of handwritten mathematical expression recognition but also opens up new avenues for research in the fields of multi-modal and multi-task learning. Moving forward, we plan to explore the scalability of our approach on larger datasets, investigate the integration of additional modalities, and further refine the model architecture to enhance its applicability in real-world scenarios.

\vspace{12pt}

\end{document}